\documentclass[journal,twoside]{IEEEtran}

\usepackage[caption=false]{subfig}
\usepackage{cite}
\usepackage{url}
\usepackage{graphicx}
\usepackage[tbtags]{amsmath}
\usepackage{graphics}
\usepackage{epsfig}
\usepackage{amsthm, amssymb}
\usepackage{mathrsfs}
\usepackage{algorithm}
\usepackage{algorithmicx}
\usepackage{algpseudocode}
\usepackage[table,xcdraw]{xcolor}
\usepackage{array}
\usepackage{tabularx}
\usepackage{xspace}
\usepackage{bbm}
\usepackage{booktabs}
\usepackage{amsthm}
\usepackage{xcolor}

\title{Energy-Efficient Split Learning for Fine-Tuning Large Language Models in Edge Networks}

\author{Zuguang~Li,~\IEEEmembership{Graduate Student Member,~IEEE},
Shaohua~Wu,~\IEEEmembership{Member,~IEEE},
Liang~Li,~\IEEEmembership{Member,~IEEE}, and
Songge~Zhang,~\IEEEmembership{Graduate Student Member,~IEEE}

\thanks{
This work was supported in part by the Peng Cheng Laboratory Major Key Project under Grants PCL2023AS1-5 and PCL2021A09-B2; and in part by the NSFC 62201071. (\textit{Corresponding author: Shaohua Wu.})

Zuguang Li is with School of Electronics and Information Engineering, Harbin Institute of Technology, Shenzhen, 518055, China, and also with the Frontier Research Center, Pengcheng Laboratory, Shenzhen, 518055, China (e-mail: lizuguang@stu.hit.edu.cn). 

Shaohua Wu is with the Guangdong Provincial Key Laboratory of Aerospace Communication and Networking Technology, Harbin Institute of Technology, Shenzhen, 518055, China (e-mail: hitwush@hit.edu.cn).

Liang Li is with the Frontier Research Center, Pengcheng Laboratory, Shenzhen, 518055, China (e-mail: lil03@pcl.ac.cn).

Songge Zhang is with the School of Electronic and Computer Engineering, Peking University, Shenzhen, 518000, China, and also with the Frontier Research Center, Pengcheng Laboratory, Shenzhen, 518055, China (email: zhangsongge@stu.pku.edu.cn).

This paper has been accepted for publication in IEEE Networking Letters.

}
}

\ifCLASSINFOpdf
\else
\fi

\begin{document}

\maketitle
\begin{abstract}
	In this letter, we propose an energy-efficient split learning (SL) framework for fine-tuning large language models (LLMs) using geo-distributed personal data at the network edge, where LLMs are split and alternately across massive mobile devices and an edge server. Considering the device heterogeneity and channel dynamics in edge networks, a \underline{C}ut l\underline{A}yer and computing \underline{R}esource \underline{D}ecision (CARD) algorithm is developed to minimize training delay and energy consumption. Simulation results demonstrate that the proposed approach reduces the average training delay and server's energy consumption by 70.8\% and 53.1\%, compared to the benchmarks, respectively.

\begin{IEEEkeywords}
Cut layer selection, fine-tuning, large language models, edge networks.
\end{IEEEkeywords}
\end{abstract}

\section{Introduction}

The rise of large language models (LLMs) marks a breakthrough in artificial intelligence (AI), with widespread applications in natural language processing, automated decision-making, and coding assistant \cite{qiu2024large}. It can be attributed to LLMs' ability to process vast amounts of unstructured data, generate context-aware responses, and perform complex problem-solving tasks across various applications. The potential of LLMs is inspiring researchers to integrate them into edge networks for offering intelligent services, e.g., resource allocation and network management \cite{friha2024llm}. Applying LLMs to edge networks can respond to dynamical network environments, enabling real-time optimization of network configurations and enhancing overall system performance.

{Parameter efficient fine-tuning (PEFT) has been explored to reduce trainable parameters in LLMs, enhancing task-specific performance in edge networks~\cite{zhang2024gradient}. By fine-tuning a small subset of parameters, PEFT achieves performance comparable to full-model fine-tuning.}
In particular, as a popular PEFT method, low-rank adaptation (LoRA) keeps the pre-trained model weights fixed and introduces trainable low-rank matrices within each transformer layer.
{In edge networks, many private institutions or companies face limitations in accessing the data stored on edge devices due to concerns about data privacy and security. This issue is exacerbated when fine-tuning requires the transfer of data to centralized servers, raising the risk of exposure to malicious actors~\cite{sun2024knowledge}.}

{Federated learning (FL) offers a promising solution to privacy concerns by enabling decentralized fine-tuning directly on multiple owners, sharing only model updates instead of raw data~\cite{Li2021infocom}.
However, applying FL for fine-tuning LLMs in edge networks remains challenging, as the process of fine-tuning the entire LLM is performed on the participating devices. This approach requires enormous computational overhead and memory, which hinders its feasibility on edge devices~\cite{mohamadi2023chatgpt}. For example, fine-tuning a T5-Large model with LoRA requires memory of 7.1 GB in a mini-batch training, while a Jetson Nano, an edge AI platform, has only 4 GB of RAM~\cite{ouyang2024pluto}.}

To address this issue, {split learning (SL) offers an alternative to reduce the computational burden on edge devices while preserving data privacy~\cite{wu2023split}. In SL, the initial layers of a model are processed locally on the device, and only the intermediate activations (not the raw data) are sent to the server, which then completes the remaining layers of the model~\cite{wu2024device}.
In~\cite{liu2024resource}, an SL-based fine-tuning approach is introduced, where the devices fine-tune the initial layers while servers handle the rest. Similarly, a split LoRA-based fine-tuning framework is presented, where the embedding and task modules are executed on devices, with the encoder kept on the server \cite{wang2024federated}. Most existing literature focuses on static split strategies and neglects server-side energy consumption.}

However,  {optimizing LLMs in edge networks is challenging in achieving the best possible latency and energy efficiency performance. \textit{Firstly},
using a static cut is suboptimal due to heterogeneous devices and the dynamic channel in edge networks. 
\textit{Secondly}, static server resource configurations can lead to inefficient energy use as workloads vary with the cut layer.} Therefore, an adaptive strategy that dynamically adjusts the cut layer and server's computational resource based on real-time conditions is necessary to optimize both latency and energy efficiency.

In this letter, \textit{firstly}, we propose an energy-efficient SL framework for fine-tuning LLMs, which aims to enhance training and energy efficiency in edge networks. \textit{Secondly}, we analyze the factors of the training delay and energy consumption in detail, then formulate an optimization problem to minimize both the training delay and the server's energy consumption. \textit{Thirdly}, to solve this problem, we design an algorithm of \underline{C}ut l\underline{A}yer and computing \underline{R}esource \underline{D}ecision, named CARD. It decomposes this problem into independent sub-problems and then jointly determines the cut layer selection and server's computation resource allocation. \textit{Finally}, extensive simulations demonstrate the performance of the proposed approach through a physical platform with 5 NVIDIA Jetson edge devices. 
The main contributions of this work are summarized as follows:
\begin{itemize}
    \item We propose an energy-efficient SL framework for fine-tuning LLMs, which enhances the training and energy efficiency in edge networks.
 
    \item We design the CARD algorithm to determine the cut layer and server's computing resource allocation.

\end{itemize}

The remainder of this letter is organized as follows. Section~\ref{sec: system model} outlines the proposed framework. Section~\ref{sec: Delay and Energy Analysis} analyzes the delay and energy consumption. Section~\ref{sec: online algorithm} describes the CARD algorithm.  Section~\ref{sec: simulation} presents the simulation results. Finally, Section~\ref{sec: conclusion} concludes this letter. 

\section{Proposed Framework} \label{sec: system model}

\subsection{LoRA-Based LLM Fine-Tuning}

\begin{figure}[t]
    \vspace{-0.3cm}
	\renewcommand{\figurename}{Fig.}
	\centering
	\includegraphics[width=0.43\textwidth]{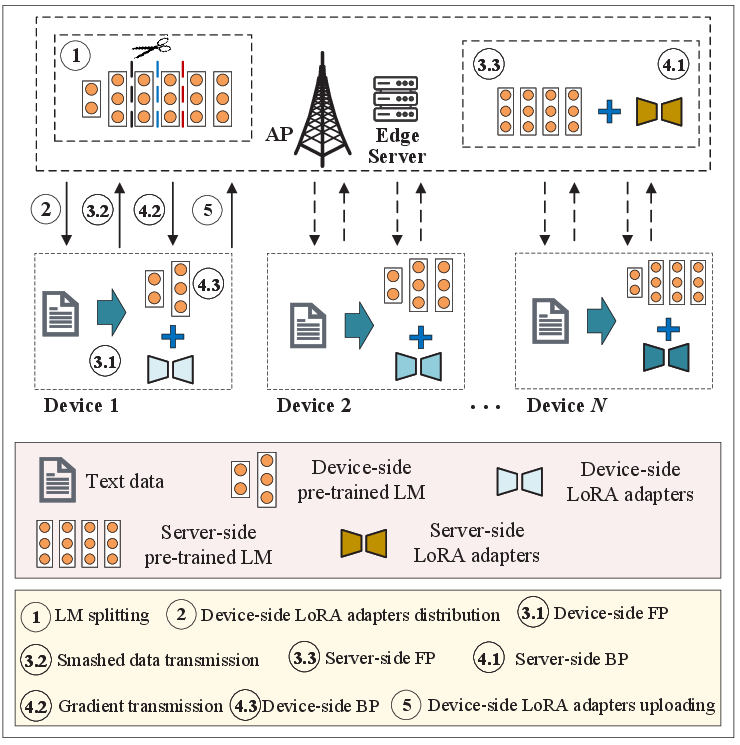}
	\caption{The proposed SL framework for fine-tuning LLMs.}
	\label{fig:system_model}
    \vspace{-0.4cm}
\end{figure}

A typical LoRA-based LLM fine-tuning scenario is considered, in which multiple edge devices collaborate with an access point (AP) to fine-tune a shared pre-trained LLM. Each device trains only the initial layers of the LLM, reducing local computation and memory requirements. The AP with an edge server coordinates to complete the remaining training process with each device, as depicted in Fig.~\ref{fig:system_model}. This approach is ideal for resource-constrained environments, enabling collaborative fine-tuning while enhancing training and energy efficiency.

\begin{figure}[t]
    \vspace{-0.3cm}
	\renewcommand{\figurename}{Fig.}
	\centering
	\includegraphics[width=0.33\textwidth]{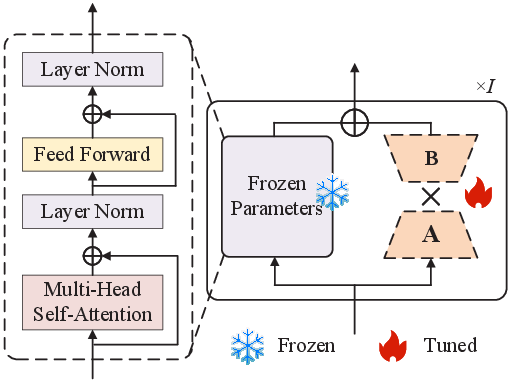}
	\caption{LoRA-based fine-tuning.}
	\label{fig: LoRA adapter}
    \vspace{-0.4cm}
\end{figure}

Let $\textbf{W} \in \mathbb{R}^{P \times Q}$ denote a pre-trained weight matrix of the LLM, with its corresponding model update expressed as $\textbf{W} + \Delta \textbf{W} = \textbf{W} + \textbf{A}\textbf{B}$. Here, $\textbf{A} \in \mathbb{R}^{P \times Z}$ and $\textbf{B} \in \mathbb{R}^{Z \times Q}$ are a pair of decomposed low-rank matrices, and $Z \ll \min(P, Q)$, as shown in Fig.~\ref{fig: LoRA adapter}. 
In the LoRA-based fine-tuning process, the pair of decomposed low-rank matrices are used for adaptation, while the weight matrix $\textbf{W}$ is frozen. 
The edge device is indexed by $ m \in \mathcal{M}=\{1, 2, ..., M\}$, and its local dataset is denoted by $\mathcal{D}_m = \{\textbf{x}_{m,k}, y_{m,k}\}_{k =1}^{D_m}$. Here, $\textbf{x}_{m,k}$ and $y_{m,k}$ are the $k$-th input data and its corresponding label. A device-side pre-trained model and its associated set of trainable LoRA adapters are denoted by $\textbf{W}_{m}^D$ and $\textbf{R}_{m}^D$, respectively. $\textbf{R}_{m}^D = \{\textbf{A}_{m}^{1}, \textbf{B}_{m}^{1}, \dots, \textbf{A}_{m}^{c_{m}}, \textbf{B}_{m}^{c_{m}}\}$, where $\textbf{A}_{m}^{i}$ and $\textbf{B}_{m}^{i}$ are the pair of low-rank matrices of $i$-th LoRA adapter, and $c_{m}$ is the cut layer. 

For the edge server deployed in the AP, a server-side pre-trained model and its corresponding set of trainable LoRA adapters are denoted by $\textbf{W}_{m}^S$ and $\textbf{R}_{m}^S$, respectively. Similarly, $\textbf{R}_{m}^S = \{\textbf{A}_{m}^{c_{m} + 1}, \textbf{B}_{m}^{c_{m} + 1}, \dots, \textbf{A}_{m}^{I}, \textbf{B}_{m}^{I} \}$, where $I$ is the total number of transform layers in the LLM. 
The global pre-trained LLM is denoted by $\textbf{W} = \{\textbf{W}_{m}^D; \textbf{W}_{m}^S \}, ~\forall m\in \mathcal{M}$. The objective of split LoRA-based fine-tuning is to find the optimal LoRA adapter configuration that minimizes the global loss across all participating devices, i.e., 
\begin{equation}  \label{eq: global loss function}
 \underset{\textbf{R}_{m}^D, \textbf{R}_{m}^S} {\text{min} } \frac{\sum_{m \in \mathcal{M}} |\mathcal{D}_m| L_m(\textbf{W} | \textbf{R}_{m}^D, \textbf{R}_{m}^S) }{\sum_{m \in \mathcal{M}} |\mathcal{D}_m|}.
\end{equation}
Here, $L_m(\textbf{W} | \textbf{R}_{m}^D, \textbf{R}_{m}^S)$ is the local loss function for device $m$ over local dataset $\mathcal{D}_m$.

\subsection{Proposed SL Framework for fine-tuning LLMs}

In this section, the proposed SL framework for fine-tuning LLMs is presented in detail. { The basic idea is to leverage the SL method, allowing devices to fine-tune the first layers of the LLM while the server handles the remaining layers. Before the initiation of fine-tuning, the parameters in LoRA adapters are initialized randomly.} After that, the LLM is executed between devices and the server in consecutive training rounds until the optimal LoRA adapters are identified. The training round is indexed by $n \in \mathcal{N}=\{1,2,..., N\}$, and the detailed workflow of the proposed framework is as follows. 

\textit{Stage 1 - LLM splitting}. The edge server selects device $m$ to execute LLM fine-tuning collectively. Then, it splits the trainable LoRA adapters into device-side LoRA adapters $\textbf{R}_{m,n}^{D}$ and server-side LoRA adapters $\textbf{R}_{m,n}^{S}$ based on cut layer $c_{m,n}$. 

\textit{Stage 2 - Device-side LoRA adapters distribution}. The edge server transmits the device-side LoRA adapters and the cut layer index to device $m$ over wireless channels. At training round $n$, device $m$ collaborates with the server to perform the LLM fine-tuning for $T_{m,n}$ local epochs, indexed by $t_{m,n} \in \mathcal{T}_{m,n} = \{1, 2, . . . , T_{m,n}\}$. For simplicity, $T_{m,n}$ and $t_{m,n}$ are represented as $T$ and $t$, respectively.

\textit{Stage 3 - Pre-trained model forward propagation (FP)}. The input data hosted on device $m$ is passed through the LLM's layers to generate predicted results, i.e., the FP process. 
It consists of three steps as follows.

\textit{ $\bullet$ Device-side FP:} Device $m$ randomly draws a mini-batch of data samples $\mathcal{H}_{m,n}(t) \subseteq \mathcal{D}_m$ for fine-tuning the device-side pre-trained model $\textbf{w}_{m,n}^D$ locally. 
Let $\textbf{x}_{m,n}^{t}$ and $y_{m,n}^{t}$ represent the input data and its label for a mini-batch at local epoch $t$, respectively. The set of trainable LoRA adapter for the device-side pre-trained model is defined as $\textbf{R}_{m, n}^{D,t} = \{\textbf{A}_{m,n}^{1, t}, \textbf{B}_{m,n}^{1, t}, \dots, \textbf{A}_{m,n}^{c_{m,n}, t}, \textbf{B}_{m,n}^{c_{m,n}, t}\}$. {Device $m$ feeds a mini-batch of data into the device-side pre-trained model, producing the smashed data at the cut layer.} The smashed data from device $m$ is denoted by
\begin{equation}  \label{eq: smashed data}
  \begin{split}
    \textbf{S}_{m,n}(t) = h \big( \textbf{w}_{m,n}^D | &\textbf{R}_{m, n}^{D,t},~ \textbf{x}_{m,n}^{t}   \big),~
    \forall m,~ n \in \mathcal{M}, ~\mathcal{N}.
  \end{split}
\end{equation}
Here, $h \big(\textbf{w} | \textbf{R}, \textbf{x}\big)$ represents the mapping between input data $\textbf{x}$ and the output, given the model parameter $\textbf{w}$ and the trainable device-side LoRA adapters $\textbf{R}$. 

\textit{$\bullet$ Smashed data transmission:} Device $m$ transmits its smashed data and corresponding label to the server over wireless channels after completing the device-side model's FP.

\textit{$\bullet$ Server-side FP:} The server feeds the smashed data into the server-side model for executing the server-side model's FP process. Let $\textbf{R}_{m, n}^{S,t} = \{ \textbf{A}_{m,n}^{c_{m,n} + 1, t}, \textbf{A}_{m,n}^{c_{m,n} + 1, t}, \dots, \textbf{A}_{m,n}^{I, t}, \textbf{B}_{m,n}^{I, t}\}$ denote the set of trainable LoRA adapter of the server-side pre-trained model at local epoch $t$. 
Therefore, the predicted result of the pre-trained LLM is denoted by
  \begin{equation}  \label{eq: predicted result}
  \begin{split}
    \hat{\textbf{y}}_{m,n}(t) = h \big( \textbf{w}_{m,n}^S | &\textbf{R}_{m, n}^{S,t},~  \textbf{S}_{m,n}  \big),~
    \forall m,~ n \in \mathcal{M}, ~\mathcal{N}.
  \end{split}
  \end{equation}

\textit{Stage 4 - Pre-trained model backward propagation (BP)}. The LoRA adapters are updated by minimizing the loss function, i.e., the BP process. It consists of three steps as follows.

\textit{$\bullet$ Server-side BP:} By comparing the corresponding labels and predicting results, the server
gets the gradients of the decomposition matrices $\textbf{A}$ and $\textbf{B}$ of $i$-th LoRA adapter, denoted by $\textbf{G}_{m,n}^{i, t}$ and $\tilde{\textbf{G}}_{m,n}^{i, t}$, respectively. Then, the $i$-th server-side LoRA adapter is updated through
\begin{equation}  \label{eq: server-side LoRA adapters}
  \begin{split}
    & \textbf{A}_{m,n}^{i, t} \leftarrow \textbf{A}_{m,n}^{i, t - 1} - \gamma_S \textbf{G}_{m,n}^{i, t}, \\
    & \textbf{B}_{m,n}^{i, t} \leftarrow \textbf{B}_{m,n}^{i, t - 1} - \gamma_S \tilde{\textbf{G}}_{m,n}^{i, t}. 
  \end{split}
\end{equation}
Here, $\gamma_S$ is the server-side learning rate. 

\textit{$\bullet$ Gradient transmission:} The server transmits the smashed data's gradient to device $m$.

\textit{$\bullet$ Device-side BP:}  The device fine-tunes its device-side pre-trained model after receiving the smashed data's gradient. The $j$-th device-side LoRA adapter is updated through
\begin{equation}  \label{eq: device-side LoRA adapters}
  \begin{split}
    & \textbf{A}_{m,n}^{j, t} \leftarrow \textbf{A}_{m,n}^{j, t - 1} - \gamma_m \textbf{G}_{m,n}^{j, t}, \\
    & \textbf{B}_{m,n}^{j, t} \leftarrow \textbf{B}_{m,n}^{j, t - 1} - \gamma_m \tilde{\textbf{G}}_{m,n}^{j, t}. 
  \end{split}
\end{equation}
Here, $\gamma_m$ is the learning rate of device $m$. Note that one local epoch for fine-tuning the pre-trained LLM is completed with Steps 3 and 4. 

\textit{Stage 5 - Device-side LoRA adapters uploading}. Device $m$ sends its device-side LoRA adapters $\textbf{R}_{m, n}^{D,T}$ to the server over wireless channels after $T$ local epochs.
The server updates the whole trainable LoRA adapters $\textbf{R}_{m, n}$ as follows
\begin{equation}  \label{eq: device-side LoRA adapters uploading}
  \begin{split}
    \textbf{R}_{m, n} = \{\textbf{R}_{m, n}^{D,T}; \textbf{R}_{m, n}^{S,T}\}.
  \end{split}
\end{equation}
Stages 1-5 are repeated for all the participating devices until a satisfactory model performance is achieved.

\section{System Model} \label{sec: Delay and Energy Analysis}

\subsection{LLM Fine-tuning Delay}

\subsubsection{Computation Delay}
In each training round, the device performs fine-tuning for the device-side pre-trained model, whose computation delay is given by
\begin{equation}\label{equ: Device-side model computation delay}
		d^{D,C}_{m,n} = \frac{\eta_D\left(c_{m,n}\right)}{f_m^D\delta_m^D\sigma_m^D}.
\end{equation}
Here, $\eta_D\left(c_{m,n}\right)$ is the total floating point operations (FLOPs) of the device-side pre-trained model at cut layer $c_{m,n}$. $f_m^D$ is the graphics processing unit (GPU) core frequency of device $m$, $\delta_m^D$ is the number of FLOPs per core of the GPU in a processing unit cycle, and $\sigma_m^D$ is the number of cores of device $m$'s GPU~\cite{liu2024resource}. We assume that $f_m^D$, $\delta_m^D$, and $\sigma_m^D$ are constants in respect of device $m$.
Similarly, the computation delay of the server is given by
\begin{equation}\label{equ: Server-side model computation delay}
		d^{S,C}_{m,n} = \frac{\eta - \eta_D\left(c_{m,n}\right)}{f_{m,n}^S\delta^S\sigma^S},
\end{equation}
where $\eta$ is the total FLOPs of the pre-trained LLM. $\delta^S$, and $\sigma^S$ are constants. $f_{m,n}^S$ is the server's GPU frequency, which is an adjustable variable to meet the energy consumption constraint of the server. 

\subsubsection{Transmission Delay}
During the fine-tuning process, necessary data, such as LoRA adapters, smashed data, and its gradients, are exchanged between the device and server, incurring transmission delay. The total transmission delay for a training round can be given by 
\begin{equation}\label{equ: transmission delay}
	\begin{split}
		D_{m,n}^{V} & = T_{m, n} \bigg( \frac{ \varphi S (c_{m,n})}{R^{D}_{m,n}} + \frac{ \varphi \tilde{S} (c_{m,n})}{R^{S}_{m,n}} \bigg) + \frac{A (c_{m,n})}{R^{D}_{m,n}} \\
        & + \frac{A (c_{m,n})}{R^{S}_{m,n}}, ~ \forall m, ~n \in \mathcal{M},~ \mathcal{N}.
	\end{split}
\end{equation}
Here, $\varphi$ is the compression ratio for smashed data and gradient. $S (c_{m,n}) $ and $\tilde{S} (c_{m,n})$ are the data size of smashed data and gradient at cut layer $c_{m,n}$, respectively. $R^{D}_{m,n}$ and $R^{S}_{m,n}$ are the transmission rates at uplink and downlink, respectively. {The transmission rate is converted from signal-to-noise ratio (SNR) by the channel quality indication (CQI) to modulation and coding scheme (MCS) mapping table \cite{3gpp2022phsical}, i.e., $R_{m,n} =  B_{m,n} y(\text{SNR}_{m,n})$. Here, $B_{m,n}$ is the bandwidth allocated for the $m$-th device, and $y(\cdot)$ is the conversion function mapping SNR to the spectral efficiency.}
$A (c_{m,n})$ is the data size of the device-side LoRA adapters.

\subsubsection{Overall Training Delay}
Considering all the components of computation and communication delays, the model training time for device 
$m$ and the server during a training round can be expressed as
\begin{equation}\label{equ:model training delay}
	\begin{split}
		D_{m,n} =  D_{m,n}^{C} + D_{m,n}^{V}, ~\forall m,~ n \in \mathcal{M},~ \mathcal{N}.
	\end{split}
\end{equation}
Here, $D_{m,n}^{C}$ is the total computation delay, defined as $D_{m,n}^{C} = T_{m, n} \left(d^{D,C}_{m,n} + d^{S,C}_{m,n} \right) $.

\subsection{Energy Consumption}
In the split fine-tuning framework, the server needs to work with all participating devices to complete the LLM fine-tuning, which imposes a heavy computation-related energy consumption burden. {The power consumption and frequency relationship is cubic for the LLM fine-tuning on GPU,} i.e., $P^{S,C}_{m,n} = \xi (f_{m,n}^S)^3 $. Here, $\xi$ is a coefficient representing power consumption per cubic cycle per second (in $\text{Watt}/(\text{cycle/s})^3$), varying based on the GPU architecture.
Hence, the computational energy consumption of the server can be defined as $E_{m,n} = T_{m, n} d_{m,n}^{S,C} P^{S,C}_{m,n}$, i.e., 
\begin{equation}
    \label{eq: computation-related energy consumption of the server}
   \begin{split}
        E_{m,n} = \frac{T_{m, n} \xi (f_{m,n}^S)^2 \left( \eta - \eta_D\left(c_{m,n}\right) \right)}{\delta^S\sigma^S}.
   \end{split}
\end{equation} 

\subsection{Problem Formulation}
The proposed scheme aims to minimize both the training delay and the server’s energy consumption. Hence, we define a cost function through weakly Pareto optimal solutions~\cite{wang2019delay}, which integrates the normalized training delay and server's computational energy consumption, i.e.,
\begin{equation}	\label{eq: cost of both training delay and energy consumption}
   \begin{split}
    U (f_{m,n}^S, c_{m,n}) = w \frac{D_{m,n} - D_{\min}}{D_{\max}- D_{\min}} + (1- w) \frac{E_{m,n}- E_{\min}}{E_{\max} - E_{\min}}.
    \end{split}
\end{equation}
Here, $0 \le w \le 1$ represents the weighting factor. $D_{\min}$ and  $D_{\max}$ are the minimum and maximum value of delay for device $m$ during training round $n$, respectively. { Moreover, $E_{\min}$ and $E_{\max}$ are the minimum and maximum value of energy consumption for device $m$ during training round $n$, respectively. $D_{\max}$ and $E_{\min}$ can be obtained when $c_{m,n} = I$ and $f_{m,n}^S = F_{\min}^{m,S}$. $D_{\min}$ and $E_{\max}$ can be obtained when $c_{m,n} = 0$ and $f_{m,n}^S = F_{\max}^{S}$. 
The optimal problem can be formulated as:}
\begin{equation}	\label{eq: Problem 1}
   \begin{split}
	  {\mathbf{P}_1:\;}  &\underset{ \{f_{m,n}^S, c_{m,n}\}_{\substack{m\in \mathcal{M}, \\ ~n \in \mathcal{N}}}}
     {\text{min} }  \sum_{m \in \mathcal{M}} \sum_{n \in \mathcal{N}} U (f_{m,n}^S, c_{m,n})  \\
        &~~~~~~\text{s.t.} ~ c_{m,n} \in \{0, 1, \cdots,I\}, ~\forall m,~ n \in \mathcal{M},~ \mathcal{N}, \\
     &~~~~~~~~~~ F_{\min}^{m,S} \le f_{m,n}^S \le F_{\max}^S, ~\forall m,~ n \in \mathcal{M},~ \mathcal{N}.
    \end{split}
\end{equation}
{Here, the first constraint guarantees the feasibility of the cut layer selection. The secondary constraint requires that the server's GPU frequency does not surpass the upper limit $F_{\max}^S$, but is not less than the lower limit $F_{\min}^{m,S}$. 
Here, $F_{\max}^S$ is the maximum frequency the server's can reach, and $ F_{\min}^{m,S} = f_m^D\delta_m^D\sigma_m^D/(\delta^S\sigma^S), \forall m$, which stems from the assumption that the computational power of the server is more powerful than that of devices to ensure the effectiveness of split learning.}

The problem $\mathbf{P}_1$ is a mixed integer nonlinear programming problem due to the inclusion of continuous and integer variables. The proposed algorithm for solving this problem is discussed in the next section.

\section{CARD Algorithm} \label{sec: online algorithm}

\begin{algorithm}[t] 
	\caption{{CARD algorithm}}
	\label{ag: CARD algorithm}
	\begin{algorithmic}[1]
        \State Compute $(f_{m,n}^{S})^\star$ based on Eq.~\eqref{eq: optimal GPU frequency};
        \State Set $U_{\min}$;
        \For{$ c_{m,n} = 0$ to $I$}
        \State Compute $U$ based on Eq.~\eqref{eq: cost of both training delay and energy consumption};
        \If{$U < U_{\min}$}
        \State  $U_{\min} = U $;
        \State $c_{m,n}^{\star} = c_{m,n}$;
        \EndIf
        \EndFor
        \State \Return $c_{m,n}^{\star}$ and $(f_{m,n}^{S})^\star$.
	\end{algorithmic}
    \vspace{0.1cm}
\end{algorithm}

Considering the independence among these devices, we decompose problem $\mathbf{P}_1$ into multiple sub-problems to allow for efficient solutions.
Each sub-problem is to jointly determine the optimal cut layer and computing resource allocation for device $m$ at training round $n$, which can be expressed as follows:
\begin{equation}	\label{eq: Problem 2}
   \begin{split}
	 {\mathbf{P}_2:\;} & \underset{ f_{m,n}^S, c_{m,n}} {\text{min} }~   U (f_{m,n}^S, c_{m,n})  \\
     &~\text{s.t.} ~ c_{m,n} \in \{0, 1, \cdots,I\}, ~\forall m, ~n \in \mathcal{M}, ~\mathcal{N}, \\
     &~~~~~ F_{\min}^{m,S} \le f_{m,n}^S \le F_{\max}^S, ~\forall m, ~n \in \mathcal{M},~ \mathcal{N}.
    \end{split}
\end{equation}
To handle the problem $\mathbf{P}_2$, we decompose it into two disjoint subproblems, i.e., lower-layer and upper-layer problems, to determine the optimal cut layer and computing resource allocation.

\subsubsection{Upper-Layer Problem} 
Given the cut layer decision, the problem $\mathbf{P}_2$ can be decomposed into an upper-layer problem, which is formulated as:
\begin{equation}	\label{pb: upper-layer problem}
	\begin{split}
		{\mathbf{P}_3:\;}	&  \underset{f_{m,n}^{S} } {\text{min} } ~   U (f_{m,n}^S) \\
		& \text{s.t.}~  F_{\min}^{m,S} \le f_{m,n}^S \le F_{\max}^S, ~\forall m, ~n \in \mathcal{M},~ \mathcal{N}.
	\end{split}
\end{equation} 
By analyzing the second derivative of the function of the problem $\mathbf{P}_3$, we can get $ {\partial}^2 U(f_{m,n}^{S}) / {\partial} (f_{m,n}^{S})^2 > 0 $. Hence, the function $U (f_{m,n}^{S})$ is convex. Then, by analyzing the first derivative of the function, the optimal solution for the server's GPU frequency is given by:
\begin{equation}
	(f_{m,n}^{S})^\star =
	 \begin{cases}
		F_{\min}^{m,S},  \hfill  ~~ \text{if} ~ Q < F_{\min}^{m,S}, \\
        Q, \hfill ~~~~~~~ ~~ ~~ \text{if} ~ F_{\min}^{m,S} \le Q \le F_{\max}^S, \\
		F_{\max}^S, \hfill  ~~ \text{if} ~ F_{\max}^S < Q ,
	\end{cases}
 \label{eq: optimal GPU frequency}
\end{equation}
where $Q = \sqrt[3]{\frac{w (E_{\max} - E_{\min}) }{2 \xi (1-w)(D_{\max} - D_{\min})  }}$.

\subsubsection{Lower-Layer Problem} 
Given the server's GPU frequency decision, we formulate an upper-layer problem as follows:
\begin{equation}	\label{pb: lower-layer problem}
	\begin{split}
		{\mathbf{P}_4:\;}	& \underset{c_{m,n} } {\text{min} } ~    U (c_{m,n}) \\
		& \text{s.t.}~  c_{m,n} \in \{0, 1, \cdots,I\}, ~\forall m,~ n \in \mathcal{M},~ \mathcal{N}.
	\end{split}
\end{equation} 
The function in the problem $\mathbf{P}_4$ is non-convex because the data sizes (e.g., device-side LoRA adapter, smashed data, and its gradient) are arbitrary functions related to the cut layer. {Due to the finite number of transformer layers in an LLM, the optimal cut layer can be found by the brute-force approach. The CARD algorithm is summarized in Alg.~\ref{ag: CARD algorithm}. Its time complexity is $O(I)$, which requires $I+1$ iterations to find the optimal cut layer and the server's GPU frequency.}

\section{Simulation Results} \label{sec: simulation}

\subsection{Simulation Setup}

\begin{table}[t] 
	\footnotesize
    \renewcommand\arraystretch{1.5}
		\vspace{-0.3cm}
	\centering
	\caption{The Server and Devices Settings.}
	\label{Table: The Server and Devices Settings}
			\vspace{-0.2cm}
	\begin{tabular}{c|c|c|c}
		\hline
		\hline
	  \textbf{Type} & \textbf{Platform} & \textbf{GPU Max Frequency} & \textbf{Cores}\\
		\hline
        Server &  Nvidia RTX 4060Ti& 2.46 GHz & 3072 \\
        \hline
	  Device 1 & Jetson AGX Orin & 1.3 GHz & 2048  \\
        \hline
        Device 2 & Jetson AGX Orin & 1.0 GHz & 2048  \\
        \hline
        Device 3 & Jetson AGX Orin & 0.7 GHz & 1792  \\
        \hline
        Device 4 & Jetson Orin NX & 0.7 GHz & 1024  \\
        \hline
        Device 5 & Jetson AGX Nano & 0.5 GHz & 512  \\
        \hline
	\end{tabular}
	\vspace{-0.3cm}
\end{table}	

\begin{table}[t]
	\footnotesize
    \renewcommand\arraystretch{1.5}
		\vspace{-0.3cm}
	\centering
	\caption{Simulation parameters.}
	\label{Table: Simulation parameters}
			\vspace{-0.2cm}
	\begin{tabular}{cc|cc}
		\hline
        \hline
	  \textbf{Parameter} & \textbf{Value}  &  \textbf{Parameter} & \textbf{Value} \\
        \hline
        $\delta_m^D$  & 2 & $\delta^S$ & 2 \\
        $\xi$  & $10^{-25}$ &  $w$  & 0.2  \\
        $T_{m, n}$  & 5 & $\varphi$ & 0.1 \\
        \hline
	\end{tabular}
	\vspace{-0.1cm}
\end{table}	

We employ a 1B LLaMA 3.2 model with 32-layer transformer decoders \cite{liu2024spinquant}, and consider 5 mobile devices and one AP with an edge server. 
These devices within the coverage area of the AP request to access and execute the LLM fine-tuning task at the beginning of each experimental trial. The sets of the server and devices are as shown in Table~\ref{Table: The Server and Devices Settings}. Here, the computational capabilities of the devices gradually decrease from Device 1 to Device 5. The other main simulation parameters are concluded in Table~\ref{Table: Simulation parameters}.

\subsection{Simulation Results}

\begin{figure}[t]
    \vspace{-0.1cm}
	\centering
	\subfloat[Cut layer selection.]
    {
        \centering
    	\includegraphics[width=0.22\textwidth]{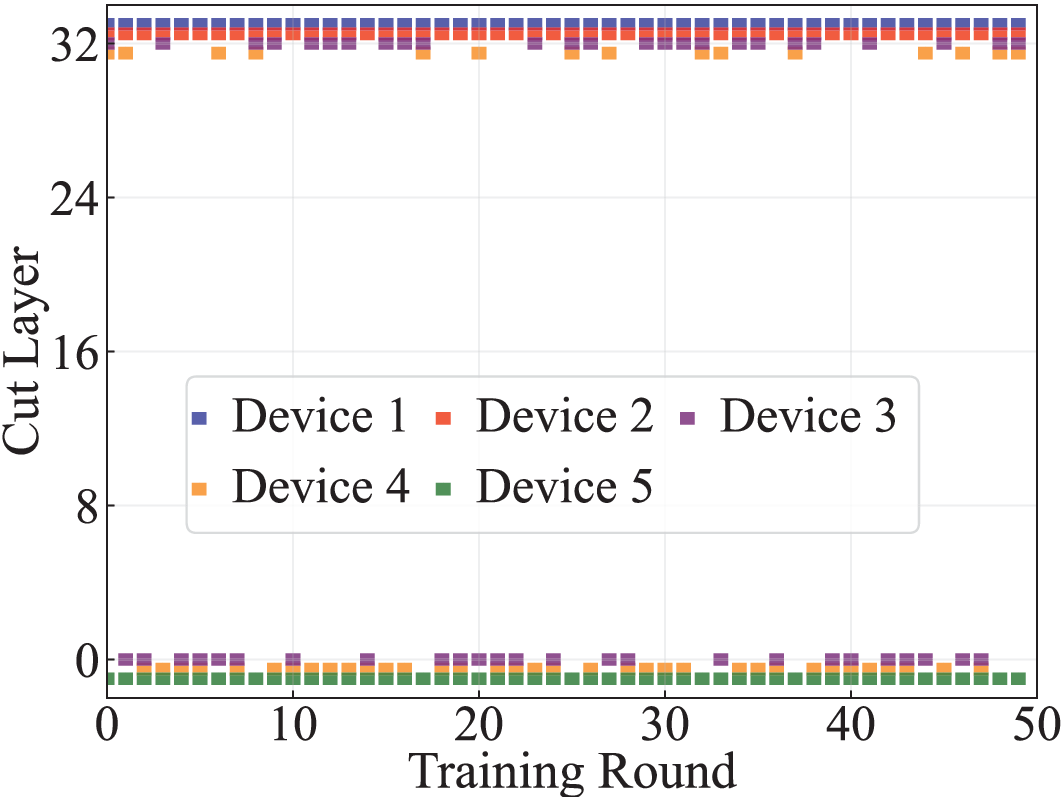}
    	\label{fig: Optimal cut layer}
    }
    \vspace{0.2cm} 
	\subfloat[Computation resource allocation.]
    {
        \centering
    	\includegraphics[width=0.22\textwidth]{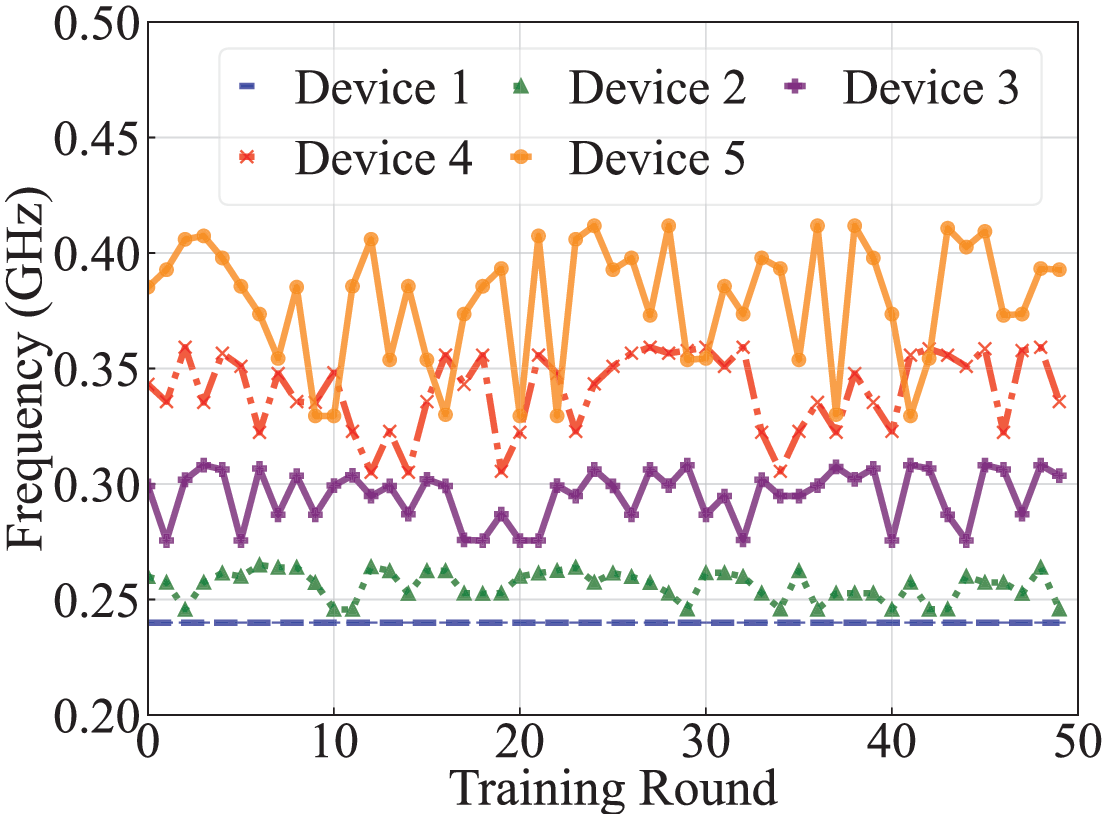}
        \label{fig: Optimal computation resource allocation}
    }
	\caption{Cut layer and computation resource allocation decisions in different training rounds.}
	\label{fig: cut layer and computation resource allocation}
 \vspace{-0.4cm}
\end{figure}

{In Fig.~\ref{fig: cut layer and computation resource allocation}, we first simulate these devices' optimal cut layer and computation resource allocation.} Fig.~\ref{fig: cut layer and computation resource allocation}(a) illustrates the optimal cut layer associated with each device, where the optimal cut layer of each device dynamically changes with training rounds due to the dynamic wireless channel. For each device, its optimal cut is either 32 or 0. This is because each transformer layer has the same computation workload and data size as the smashed data, leading to a positive correlation between the cut layer and training delay, and a negative correlation with the server's energy consumption.
In addition, as the computation power of a device decreases, i.e., from Device 1 to 5, its optimal cut layer moves from 32 to 0. Consequently, {the server allocates a higher GPU frequency to complete the LLM fine-tuning, as shown in Fig.~\ref{fig: cut layer and computation resource allocation}(b). That is because the device with lower computing power prefers to offload most of the LLM to the server rather than train them locally, reducing the training delay.}

{We then evaluate the fine-tuning performance of the proposed approach in terms of training delay and energy consumption, comparing it with two benchmarks: (\textit{i}) Server-only, where devices fine-tune the embedding module locally, and the server handles the rest. (\textit{ii}) Device-only, where devices fine-tune the embedding module and transform decoders locally, and the server handles the rest. The channel condition is set to three states, i.e., \textit{Good}, \textit{Normal}, and \textit{Poor}, corresponding to the path-loss exponent 2, 4, and 6, respectively. }

\begin{figure}[t]
    \vspace{-0.3cm}
	\centering
	\subfloat[Training delay.]
    {
        \centering
    	\includegraphics[width=0.22\textwidth]{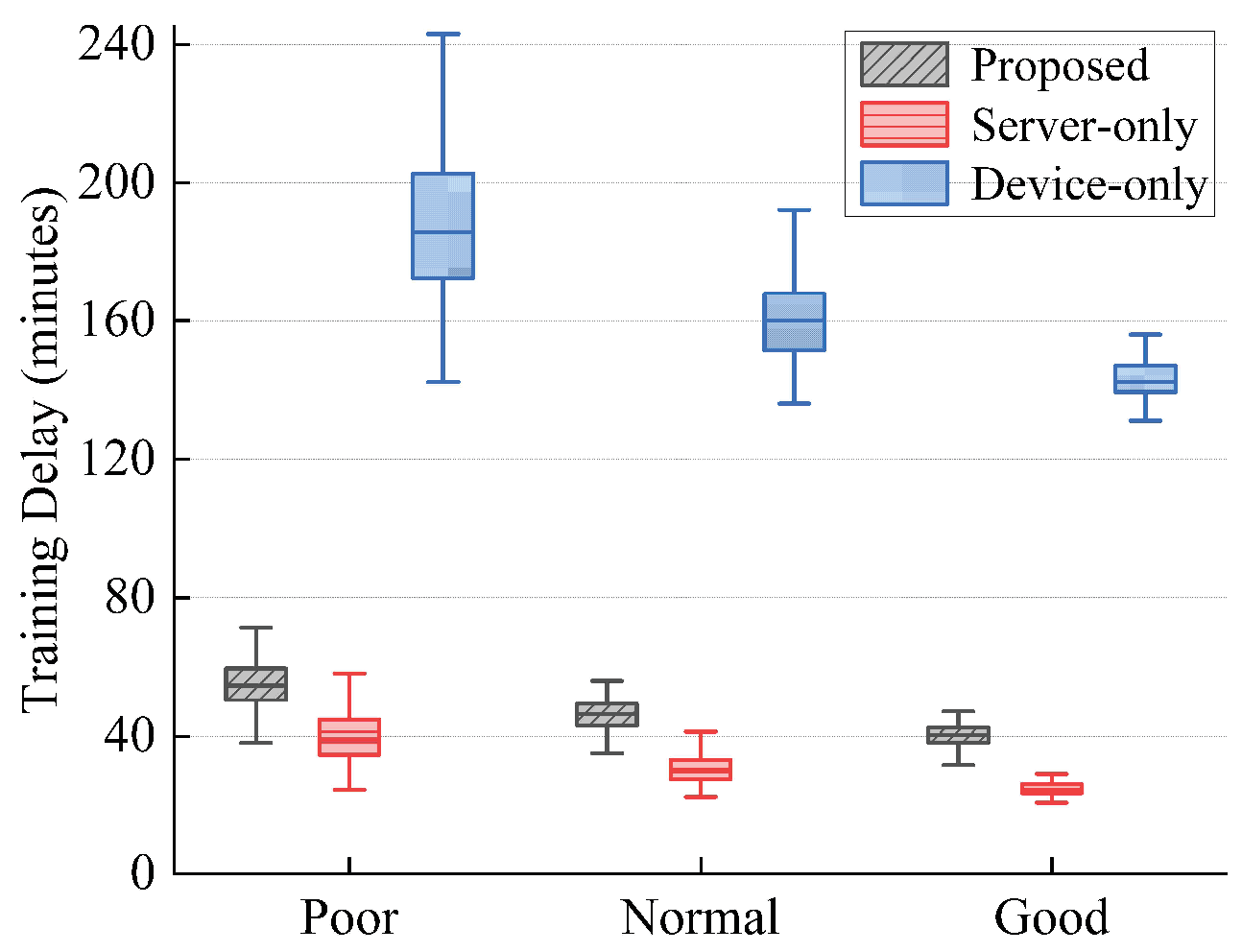}

    }
    \vspace{0.2cm} 
	\subfloat[Average computational energy consumption on the server.]
    {
        \centering
    	\includegraphics[width=0.22\textwidth]{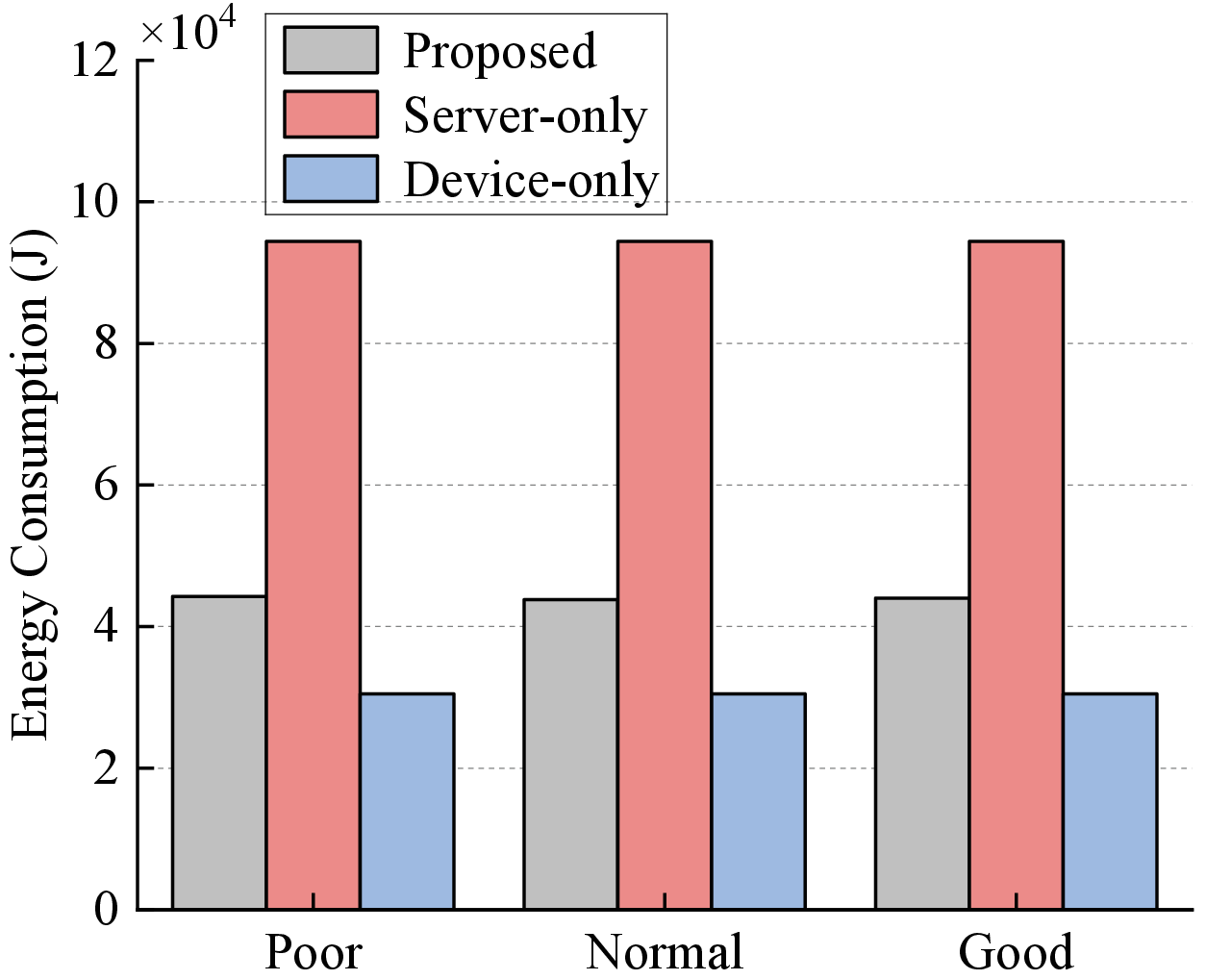}
    }
	\caption{Comparison of performance in different methods.}
	\label{fig: Average training delay and energy consumption.}
 \vspace{-0.4cm}
\end{figure}

{In Fig.~\ref{fig: Average training delay and energy consumption.}, we present the training delay and the server's average computational energy consumption during a training round for each method. Compared with the device-only method, the proposed scheme significantly reduces training delay by 70.8\%, despite introducing some computational energy consumption on the server. Additionally, while the training delay of the proposed scheme is higher than that of the server-only method, it achieves a remarkable 53.1\% reduction in the server's computational energy consumption.}

\section{Conclusion} \label{sec: conclusion}
We have proposed an energy-efficient SL framework for deploying LLMs in edge networks. We have formulated an optimization problem to achieve the trade-off between the training delay and the server’s energy consumption. To solve this problem, we have designed the CARD algorithm to jointly determine the optimal cut layer and server's computation resource allocation. Simulation results have demonstrated the effectiveness of the proposed approach. The proposed framework can be applied to accelerate LLM fine-tuning in energy-constrained edge networks. For future work, we will explore an adaptive strategy to enhance robustness against varying edge network conditions.

\bibliographystyle{IEEEtran}
\bibliography{main}
\end{document}